\documentclass[journal]{IEEEtran}
\usepackage{amsmath,amsfonts}
\usepackage{algorithmic}
\usepackage{array}
\usepackage[caption=false,font=normalsize,labelfont=sf,textfont=sf]{subfig}
\usepackage{textcomp}
\usepackage{stfloats}
\usepackage{url}
\usepackage{verbatim}
\usepackage{graphicx}
\usepackage{amsmath,amsfonts}
\usepackage{algorithmic}
\usepackage{algorithm}
\usepackage{array}
\usepackage[caption=false,font=normalsize,labelfont=sf,textfont=sf]{subfig}
\usepackage{textcomp}
\usepackage{stfloats}
\usepackage{url}
\usepackage{verbatim}
\usepackage{graphicx}
\usepackage{cite}
\usepackage{hyperref}
\usepackage{booktabs}
\usepackage{colortbl}
\usepackage{xcolor}
\usepackage{multirow}
\usepackage{float} 
\usepackage{graphicx}
\usepackage{minted}
\usepackage{amsmath, amsfonts, amssymb}
\usepackage[most]{tcolorbox}
\usepackage{lipsum} 
\usepackage{makecell}
\usepackage{color,soul}
\definecolor{modelhl}{RGB}{255,255,102}

\def\BibTeX{{\rm B\kern-.05em{\sc i\kern-.025em b}\kern-.08em
    T\kern-.1667em\lower.7ex\hbox{E}\kern-.125emX}}
\usepackage{balance}
\begin{document}
\title{Multi-Agent Inverted Transformer for Flight Trajectory Prediction}
\author{Seokbin Yoon, Keumjin Lee~\IEEEmembership{Member,~IEEE}
\thanks{Manuscript received TODO.}
\thanks{This work is supported by the Korea Agency for Infrastructure Technology Advancement (KAIA) grant funded by the Ministry of Land, Infrastructure and Transport (Grant RS-2022-00156364). (\textit{Corresponding author: Keumjin Lee)}.}
\thanks{S. Yoon and K. Lee are with the Department of Air Transport, Transportation, and Logistics, Korea Aerospace University, Goyang, South Korea. (e-mail: \{sierra.bin, keumjin.lee\}@kau.ac.kr)}}

\markboth{IEEE Transactions on Intelligent Transportation Systems}%
{How to Use the IEEEtran \LaTeX \ Templates}

\maketitle

\begin{abstract}
Flight trajectory prediction for multiple aircraft is essential and provides critical insights into how aircraft navigate within current air traffic flows. However, predicting multi-agent flight trajectories is inherently challenging. One of the major difficulties is modeling both the individual aircraft behaviors over time and the complex interactions between flights. Generating explainable prediction outcomes is also a challenge. Therefore, we propose a Multi-Agent Inverted Transformer, MAIFormer, as a novel neural architecture that predicts multi-agent flight trajectories. The proposed framework features two key attention modules: (i) masked multivariate attention, which captures spatio-temporal patterns of individual aircraft, and (ii) agent attention, which models the social patterns among multiple agents in complex air traffic scenes. We evaluated MAIFormer using a real-world automatic dependent surveillance-broadcast flight trajectory dataset from the terminal airspace of Incheon International Airport in South Korea. The experimental results show that MAIFormer achieves the best performance across multiple metrics and outperforms other methods. In addition, MAIFormer produces prediction outcomes that are interpretable from a human perspective, which improves both the transparency of the model and its practical utility in air traffic control.
\end{abstract} 


\section{Introduction}
\IEEEPARstart{T}{he} growing demand for air travel has resulted in greater airspace congestion, which necessitates advanced air traffic management (ATM) strategies, such as trajectory-based operation (TBO)~\cite{planning2007concept}. In TBO, aircraft operations are managed based on their trajectories in four-dimensional spaces (latitude, longitude, altitude, and time), which provides more precise and reliable trajectory information that air traffic controllers (ATCs) can use to optimize their decisions. In this context, accurate and reliable flight trajectory prediction is increasingly important.

Conventionally, physics-based methods for flight trajectory prediction have been widely used, which rely on kinetic models of aircraft and aerodynamic performance data such as the Base of Aircraft Data (BADA)~\cite{nuic2010user,gallo2007trajectory,dupuy2007preliminary,lymperopoulos2006model}. These approaches are effective in low-density airspace under normal conditions, but they often struggle to predict flight trajectories accurately when they deviate substantially from standard flight procedures due to interventions by ATCs~\cite{hamed2013statistical}. Therefore, various data-driven approaches have gained significant attention in recent years.

Early models based on classical multivariate regression techniques have been proposed for flight trajectory prediction~\cite{de2013machine,hong2015trajectory}. Several studies have explored the use of Gaussian mixture models (GMMs) to represent the statistical distributions of flight trajectories and apply them to flight trajectory prediction~\cite{barratt2018learning,murca2018identification,paek2020route, jung2023inferring,choi2021hybrid}. Recent advancements have increasingly focused on deep learning-based approaches. Several models for flight trajectory prediction based on long short-term memory (LSTM) networks have been proposed~\cite{shi20204,zeng2020deep,zhang2023flight}. Other studies have combined convolutional neural networks (CNNs) with LSTMs to improve the extraction of the spatio-temporal patterns of flight trajectories~\cite{ma2020hybrid,kim2021air}. Social-LSTM networks also have been proposed for multi-agent flight trajectory prediction~\cite{xu2021multi, chu2025joint}.

More recently, Transformer-based models have gained prominence due to their better capabilities in sequence representation. Vanilla Transformer architectures have been employed for long-term flight trajectory prediction~\cite{tong2023long}, and the temporal fusion Transformer has been used to improve the capture of temporal dependencies in flight trajectories~\cite{silvestre2024multi}. Other approaches have explored alternative representations of flight trajectories by applying binary encoding or inverted embedding techniques~\cite{guo2024flightbert++,yoon2025aircraft}. The agent-aware attention mechanism originally introduced in AgentFormer~\cite{yuan2021agent} has been adapted for multi-agent flight trajectory prediction~\cite{choi2023multi,deng2024multi}. 

Aircraft movements in airspace are inherently multi-agent in nature, and an effective model for flight trajectory prediction must account for both the spatio-temporal dynamics of individual aircraft and their social interactions with surrounding aircraft. Standard attention-based approaches typically rely on computing attention scores between trajectory points at different time steps for different aircraft~\cite{yuan2021agent,choi2023multi,deng2024multi}. These approaches handle spatio-temporal dynamics and social interactions jointly within a single attention layer, which inevitably increases modeling complexity and reduces the interpretability of the resulting attention scores. Several recent studies have explored hierarchical attention mechanisms to capture multi-level traffic information more efficiently, but these approaches still primarily rely on low-level time-step-to-time-step attention~\cite{yu2020spatio, sun2023mmh, xiong2023hierarchical}. Consequently, the learned attention distributions often remain irrelevant or hard to interpret from a human perspective.

Motivated by these limitations, this paper proposes a Multi-Agent Inverted Transformer (MAIFormer) for flight trajectory prediction. MAIFormer reflects the hierarchical nature of information in multi-agent flight trajectories by explicitly decoupling the modeling of intra-agent and inter-agent relationships into two dedicated attention stages. In the first stage, masked multivariate attention (MMA) is used to capture the series-wise relationships among variates for each individual aircraft trajectory. In the second stage, agent attention (AA) is applied across different flights to model social interactions between aircraft. This hierarchical design separately represents intra-agent trajectory evolution (low-level) and inter-agent influence (high-level) and is better aligned with the reasoning process of human ATCs.

\begin{figure}[t!]			
	\centering
	\includegraphics[width=\linewidth]{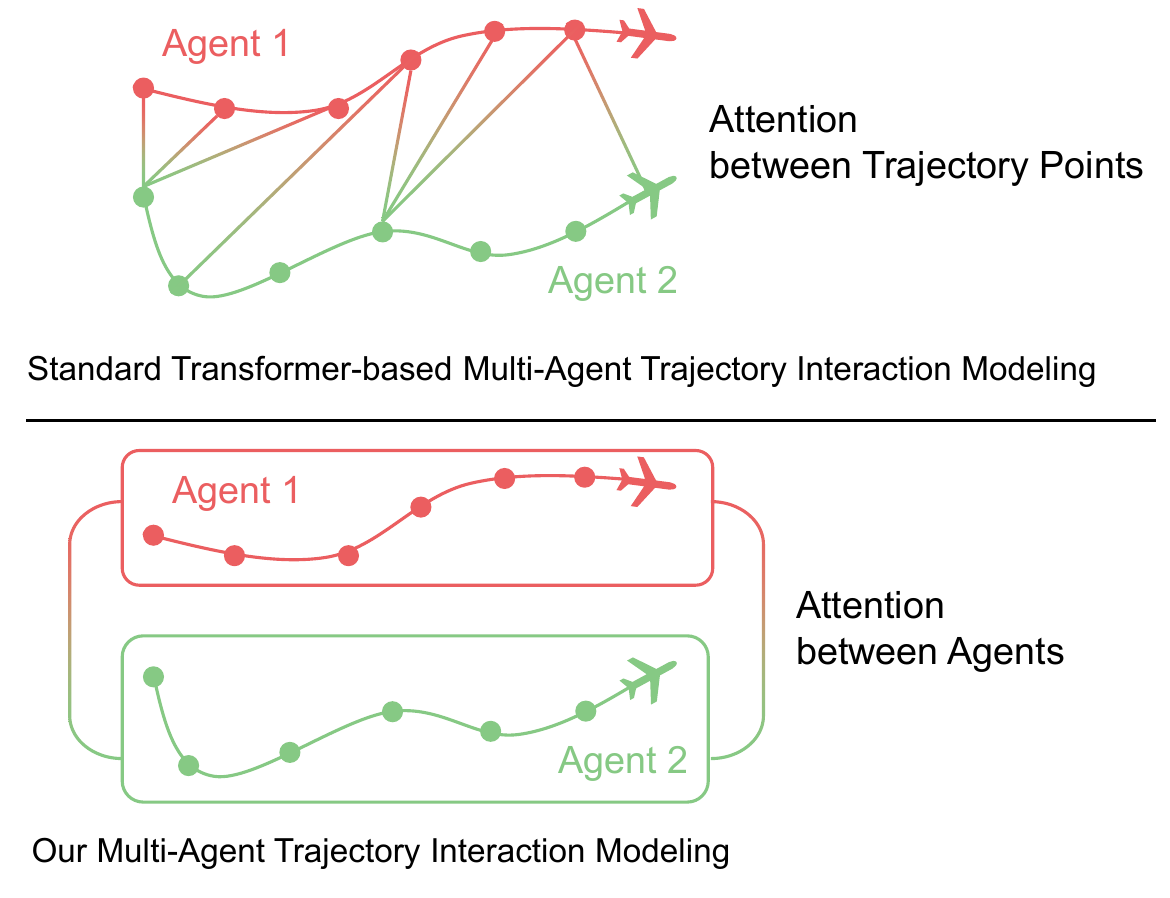}
 \vspace{-7mm} 
    \caption{Conventional approaches model inter-agent interactions by applying attention mechanisms across trajectory points, whereas MAIFormer models these interactions through direct agent-to-agent attention.}
    \label{attention_comaprison}
 \end{figure}

Furthermore, MAIFormer represents the entire trajectory of each aircraft as a single token and computes attention directly between agents (i.e., aircraft) to capture their high-level interactions, as illustrated in Figure~\ref{attention_comaprison}. This agent-level attention mechanism leads to more structured and interpretable attention distributions. Interpretability is particularly important in safety-critical domains such as ATM, where the outputs of a flight trajectory prediction model must be understandable by human operators such as ATCs.

The main contributions of this work are the following:

\begin{itemize}
    \item We propose MAIFormer, a novel multi-agent aircraft trajectory model that incorporates two specialized attention modules to reflect the hierarchical structure of multi-agent flight interactions.

    \item The proposed model demonstrates significant performance improvements in prediction accuracy over other methods when evaluated on a real-world automatic dependent surveillance-broadcast (ADS-B) flight trajectory dataset.
    
    \item MAIFormer uses a hierarchical attention mechanism to assign a single interpretable attention score to each aircraft, which enables a more intuitive understanding of how the model perceives inter-agent interactions within a traffic scene. The attention entropy analysis shows that MAIFormer yields lower attention entropy compared with AgentFormer, indicating more concentrated and thus more interpretable interaction patterns for human operators.
    
\end{itemize}

\section{Methods}\label{approach}
\subsection{Problem Formulation}
Let $\mathbf{X}$ represent the input air traffic scene involving $N$ aircraft over a time horizon $T$. The state of the \( n \)-th aircraft in the air traffic scene at time \( t \) is given by a feature vector \( \mathbf{x}_n^t \in \mathbb{R}^{1 \times F} \), where $F$ is the number of variates (e.g., latitude, longitude, altitude). The past trajectory sequence of the \( n \)-th aircraft over a time horizon \( T \) is denoted by \( \mathbf{X}_n = (\mathbf{x}_n^1, \mathbf{x}_n^2, \dots, \mathbf{x}_n^T) \in \mathbb{R}^{T \times F} \). Similarly, the future trajectory sequence of the \( n \)-th aircraft over a prediction horizon \( S \) is denoted by \( \mathbf{Y}_n = (\mathbf{y}_{n}^{1}, \mathbf{y}_{n}^{2}, \dots, \mathbf{y}_{n}^{S}) \in \mathbb{R}^{S \times F} \). Therefore, the full multi-agent past and future air traffic scenes are represented as \( \mathbf{X} = (\mathbf{X}_1, \mathbf{X}_2, \dots, \mathbf{X}_N) \in \mathbb{R}^{N \times T \times F} \) and \( \mathbf{Y} = (\mathbf{Y}_1, \mathbf{Y}_2, \dots, \mathbf{Y}_N) \in \mathbb{R}^{N \times S \times F} \), respectively. Our goal is first to train a multi-agent flight trajectory prediction model $\mathcal{F}_{\theta}(\mathbf{Y} \vert \mathbf{X})$, which minimizes the difference between $\mathbf{Y}$ and $\widehat{\mathbf{Y}}$, where $\mathcal{F}$ is the multi-agent flight trajectory prediction model, $\theta$ represents the learnable model parameters, and $\mathbf{Y}$ and $\widehat{\mathbf{Y}}$ are the actual and predicted trajectories of multi-agent aircraft, respectively. Formally, the overall learning problem is defined as:
\begin{equation}
\theta^* = \min_{\theta}  \mathcal{L}(\mathbf{Y},\widehat{\mathbf{Y}}), ~~~{\rm s.t.}~~~ \widehat{\mathbf{Y}} \sim \mathcal{F}_\theta.
\end{equation}
where $\mathcal{L}$ is a loss function that quantifies the difference between $\mathbf{Y}$ and $\widehat{\mathbf{Y}}$.

\begin{figure*}[t!]			
	\centering
	\includegraphics[width=\textwidth]{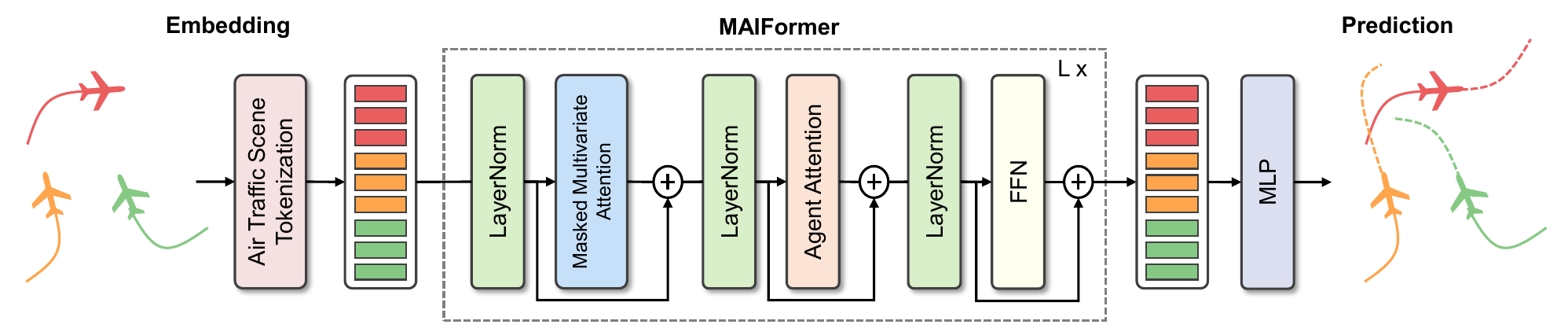}
    \vspace{-7mm}
    \caption{Overall architecture of MAIFormer. The past air traffic scene is first tokenized into variate tokens. The sequence of variate tokens from all agents is fed into stacked MAIFormer layers where spatio-temporal and social interaction patterns of the air traffic scene are captured with masked multivariate attention and agent attention, respectively.}
    \label{MAIFormer}
 \end{figure*} 
 
\subsection{Overview of the Proposed Framework}
Figure~\ref{MAIFormer} illustrates the overall architecture of the proposed method, which has three main components. First, a novel tokenization module embeds air traffic scene $\mathbf{X}$ into a sequence of variate tokens using an inverted embedding strategy~\cite{liu2023itransformer}. Second, MAIFormer, an encoder-only Transformer, extracts both the intra-agent and inter-agent relationships from the multi-agent flight trajectories within the air traffic scene. Finally, given the latent representation of $\mathbf{X}$ produced by the MAIFormer, a multi-layer perceptron (MLP) decoder is used to predict the future air traffic scene $\widehat{\mathbf{Y}}$.

MAIFormer is a fully Transformer-based encoder, and its core operation is centered around self attention (SA)~\cite{vaswani2017attention}. SA computes pairwise relationships between elements in an input sequence to determine how much attention each element should receive relative to the others. Mathematically, SA is defined as:
\begin{equation}
\text{Attention}(\mathbf{Q},\mathbf{K},\mathbf{V})= \text{Softmax}\left(\frac{\mathbf{QK}^{\top}}{\sqrt{d_k}}\right)\mathbf{V}
\end{equation}
where $\mathbf{Q}\in \mathbb{R}^{T \times D}, \mathbf{K}\in \mathbb{R}^{T \times D}$, and $\mathbf{V}\in \mathbb{R}^{T \times D}$ represent queries, keys, and values, respectively. $D$ is the model’s latent dimension, and $d_k$ is the dimensionality of the keys.

\subsection{Air Traffic Scene Tokenization}\label{atsc}
Given the multi-agent flight trajectory sequences $\mathbf{X} \in \mathbb{R}^{N\times T\times F}$, we invert and reshape it into $\mathbf{\overline{X}} \in \mathbb{R}^{(N\cdot F) \times T}$ and treat the trajectories from the multiple aircraft as a single multivariate time-series with $N\cdot F$ variates. We then embed $\mathbf{\overline{X}}$ along the time dimension as follows:
\begin{equation}
    \mathbf{C} = \text{SceneTokenization}(\mathbf{\overline{X}})
\end{equation}
$\mathbf{C}=(\mathbf{c}^1_1, \mathbf{c}^2_1, \dots, \mathbf{c}^F_1, \dots, \mathbf{c}^1_N, \dots \mathbf{c}^{F}_{N}) \in \mathbb{R}^{(N\cdot F) \times D}$ denotes the sequence of variate tokens, where each token is generated by applying a linear embedding to the original time series of each variate in $\mathbf{\overline{X}}$. Notably, the positional encoding is not applied after the tokenization, since there is no inherent ordering among agents in the air traffic scene. As a result, the input to the MAIFormer consists of $N\cdot F$ tokens, which are each a $D$-dimensional representation vector corresponding to each aircraft variate.

\begin{figure}[t!]			
	\centering
	\includegraphics[width=\linewidth]{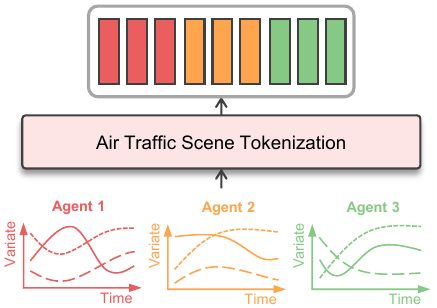}
    
    \vspace{-3mm} 
    \caption{Air traffic scene tokenization. We apply an inverted embedding to each agent’s past trajectory and generate a sequence of variate tokens from all agents. In this example, both the number of agents $N$ and the number of variates $F$ are set to three, which results in three tokens per agent and a total of nine tokens after tokenization. For consistency, we maintain these settings (three agents and three variates) in the subsequent figures.}
    \label{token}
 \end{figure} 

\subsection{MAIFormer: Multi-Agent Inverted Transformer}
MAIFormer consists of $L$ identical layers, where each layer comprises three key components: an MMA module, an AA module, and a feed-forward network (FFN) module. We define each MAIFormer layer as follows:
\begin{align}
    \mathbf{C}_{\text{ST}}^{\ell-1}   &= \text{MaskedMultivariateAttention}(\text{LN}(\mathbf{C}^{\ell-1})) + \mathbf{C}^{\ell-1}, \\
    \mathbf{C}_{\text{SC}}^{\ell-1}   &= \text{AgentAttention}(\text{LN}(\mathbf{C}_{\text{A}}^{\ell-1}))+ \mathbf{C}_{\text{A}}^{\ell-1},  \\
    \mathbf{C}^{\ell}           &= \text{FFN}(\text{LN}(\mathbf{C}_{\text{SC}}^{\ell-1}))+\mathbf{C}_{\text{SC}}^{\ell-1}, \quad \ell= (1,\ldots, L)
\end{align}
where $\mathbf{C}^{\ell} \in \mathbb{R}^{(N\cdot F) \times D}$ is the sequence of variate tokens from the $\ell$-th layer, and the output of the MMA module is $\mathbf{C}_{\text{ST}} \in \mathbb{R}^{(N\cdot F) \times D}$. The sequence of agent tokens $\mathbf{C}_{\text{A}} \in \mathbb{R}^{N \times (F \cdot D)} $ is then obtained by reshaping $\mathbf{C}_{\text{ST}}$ through agent-wise concatenation, and the output of the AA module is $\mathbf{C}_{\text{SC}} \in \mathbb{R}^{(N\cdot F) \times D}$. To improve the convergence and training stability, layer normalization (LN) was applied to the input of each module. 

The FFN in MAIFormer was designed with shared weights across all tokens, which allows the model to learn intrinsic temporal patterns across different variates. The FFN operation is defined as:
\begin{equation}
    \text{FFN}(\mathbf{C}^{\ell-1}_{\text{SC}}) = \text{GELU}(\mathbf{C}^{\ell-1}_{\text{SC}}\mathbf{W}^1 + \boldsymbol{b}^1)\mathbf{W}^2 + \boldsymbol{b}^2
\end{equation}
where $\mathbf{W}^1 \in \mathbb{R}^{D\times (4 \cdot D)}$ and $\mathbf{W}^2 \in \mathbb{R}^{(4 \cdot D) \times D}$ are learnable projection matrices, while $\boldsymbol{b}^1 \in \mathbb{R}^{4 \cdot D}$ and $\boldsymbol{b}^2 \in \mathbb{R}^{D}$ are the corresponding bias vectors. Gaussian error linear unit (GELU)~\cite{hendrycks2016gaussian} is used as the nonlinear activation function in the FFN and defined as follows: 
\begin{equation} 
\text{GELU}(x) = x \cdot \Phi(x)
\end{equation}
where $\Phi(x)$ is the cumulative distribution function (CDF) of the standard normal distribution. The pseudo-code of MAIFormer is presented in Algorithm~\ref{alg:training2}.

\begin{algorithm}[t!]
\caption{Pseudo-code of MAIFormer}\label{alg:training2}
\small

\begin{algorithmic}[1]
\REQUIRE Tokenized air traffic scene $\mathbf{C} \in \mathbb{R}^{(N \cdot F) \times D}$

\FOR{$l \gets 1$ to $L$} 
    \STATE $\mathbf{C}_{\text{ST}}^{\ell-1}   = \text{MMA}(\text{LN}(\mathbf{C}^{\ell-1})) + \mathbf{C}^{\ell-1}  $\hfill\ $\triangleright
    \mathbf{C}_{\text{ST}}^{\ell-1} \in \mathbb{R}^{(N \cdot F) \times D}$

    \STATE $\mathbf{C}_{\text{A}}^{\ell-1}   = \mathbf{C}_{\text{ST}}^{\ell-1}.\text{reshape}  $\hfill\ $\triangleright
    \mathbf{C}_{\text{A}}^{\ell-1} \in \mathbb{R}^{N \times (F \cdot D)}$

     \STATE $\mathbf{C}_{\text{SC}}^{\ell-1}   = \text{AA}(\text{LN}(\mathbf{C}_{\text{A}}^{\ell-1}))+ \mathbf{C}_{\text{A}}^{\ell-1}  $\hfill\ $\triangleright
    \mathbf{C}_{\text{SC}}^{\ell-1} \in \mathbb{R}^{N \times (F \cdot D)}$

    \STATE $\mathbf{C}_{\text{SC}}^{\ell-1}   = \mathbf{C}_{\text{SC}}^{\ell-1}.\text{reshape}  $\hfill\ $\triangleright
    \mathbf{C}_{\text{SC}}^{\ell-1} \in \mathbb{R}^{(N \cdot F) \times D}$

    \STATE $\mathbf{C}^{\ell}   = \text{FFN}(\text{LN}(\mathbf{C}_{\text{SC}}^{\ell-1})) +  \mathbf{C}_{\text{SC}}^{\ell-1} $\hfill\ $\triangleright
    \mathbf{C}^{\ell} \in \mathbb{R}^{(N \cdot F) \times D}$
\ENDFOR
\end{algorithmic}
\end{algorithm}

\subsubsection{Masked Multivariate Attention} 
\begin{figure*}[t!]			
	\centering
	\includegraphics[width=\textwidth]{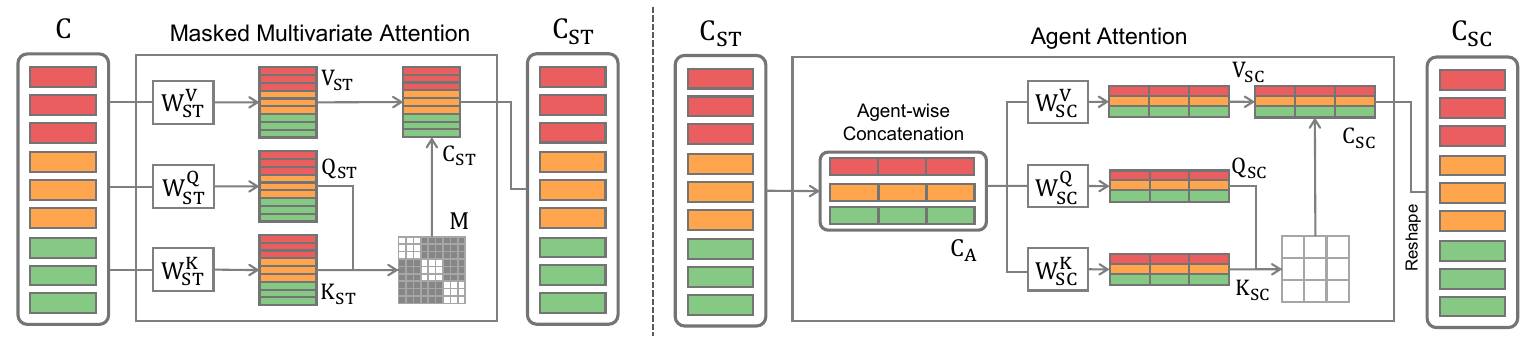}
 \vspace{-7mm}
    \caption{Illustration of masked multivariate attention (MMA) (left) and agent attention (AA) (right). In the MMA, the mask matrix $\mathbf{M}$ restricts the attention scores only to the variates from the same agent, which prevents information sharing across agents. In contrast, the AA allows all agents to share information with one another.}
    \label{attentions}
 \end{figure*}

MMA was specifically designed to model the spatio-temporal (ST) patterns within each individual flight trajectory. In MMA, when computing attention scores between variate tokens for the $i$-th agent $\mathbf{C}_i = (\mathbf{c}^1_i, \mathbf{c}^2_i, \dots, \mathbf{c}^{F}_i) \in \mathbb{R}^{F \times D}$, the variate tokens of other agents are masked using a mask matrix $\mathbf{M} \in \mathbb{R}^{(N\cdot F) \times (N\cdot F)}$, which is defined as:
\begin{equation}
\begin{aligned}
    \mathbf{M}[m, n]  = 
    \begin{cases}
        1, & \text{if} \  \left\lfloor \frac{m}{F} \right\rfloor = \left\lfloor \frac{n}{F} \right\rfloor  \\
        -\infty, & \text{otherwise}, \forall \, m, n \in \{0, 1, \dots, N \cdot F - 1\} 
    \end{cases}
\end{aligned}
\end{equation}
where $\left\lfloor \cdot \right\rfloor $ denotes the floor function. Using the mask matrix $\mathbf{M}$, we can ensure that the attentions in MMA are only computed among variates within the same aircraft, and the attentions to the variates from other aircraft are restricted. Since the mixing of variate information between different aircraft is prevented in MMA, MAIFormer more effectively captures fine-grained individual aircraft behavior, which improves prediction accuracy. The overall MMA process is illustrated in Figure~\ref{attentions} (left).

\subsubsection{Agent Attention} AA was designed to model the social context (SC) patterns between agents. To this end, SA is applied across agent tokens, which are constructed by agent-wise concatenation of the variate tokens generated from the previous MMA stage. The agent token of the $i$-th agent is defined as:
\begin{equation}
    \mathbf{C}_{\text{A},i} = (\mathbf{c}_{\text{ST},i}^{1} \oplus \mathbf{c}_{\text{ST},i}^{2} \oplus \cdots \oplus \ \mathbf{c}_{\text{ST},i}^{F}) \in \mathbb{R}^{1 \times (F \cdot D)}
\end{equation}
where $\mathbf{c}_{\text{ST},i}^{j}$ represents the $j$-th variate token of the $i$-th agent, and $\oplus$ denotes concatenation along the variate dimension. Once we obtain a whole sequence of agent tokens $\mathbf{C}_{\text{A}} \in \mathbb{R}^{N \times (F \cdot D)}$ for $N$ agents, AA is applied to that sequence as follows:
\begin{equation}
\begin{aligned}
     &\mathbf{C}_{\text{SC}} = \text{AgentAttention}(\mathbf{Q}_\text{SC},\mathbf{K}_\text{SC},\mathbf{V}_\text{SC})  \\
    &\mathbf{Q}_\text{SC}, \mathbf{K}_\text{SC}, \mathbf{V}_\text{SC} = \mathbf{C}_{\text{A}} \mathbf{W}^{\mathbf{Q}}_\text{SC}, \mathbf{C}_{\text{A}}\mathbf{W}^{\mathbf{K}}_\text{SC}, \mathbf{C}_{\text{A}} \mathbf{W}^{\mathbf{V}}_\text{SC}
\end{aligned}
\end{equation}
where each $\mathbf{W}_\text{SC} \in \mathbb{R}^{(F \cdot D) \times (F \cdot D )}$ is a learnable projection for agent tokens that transforms the agent tokens into agent queries $\mathbf{Q}_\text{SC} \in \mathbb{R}^{N \times (F \cdot D)}$, agent keys $\mathbf{K}_\text{SC} \in \mathbb{R}^{N \times (F \cdot D)}$, and agent values $\mathbf{V}_\text{SC} \in \mathbb{R}^{N \times (F \cdot D)}$.

The agent-level representations in AA allows inter-agent relationships to be effectively captured without the complexity of trajectory point-level attentions across agents, which often obscure attention interpretability. The output vector from AA is then reshaped back to its original shape, $\mathbf{C}_{\text{SC}} \in \mathbb{R}^{(N \cdot F) \times D}$, to be used in the subsequent operations in the next FFN module. The overall process of AA is illustrated in Figure~\ref{attentions} (right).

\subsubsection{MLP Decoder}
The latent representations $\mathbf{C}^L  \in \mathbb{R}^{(N \cdot F) \times D}$ from the last MAIFormer layer are passed through a MLP decoder for prediction:
\begin{equation}
    \widehat{\mathbf{Y}} = \text{MLP}(\mathbf{C}^L) 
\end{equation}
The MLP consists of multiple linear layers with GELUs, which progressively reduce the dimension from $\mathbb{R}^{D}$ to $\mathbb{R}^{S}$. The output $\widehat{\mathbf{Y}} \in \mathbb{R}^{(N \cdot F) \times S}$ is then reshaped and transposed to $\mathbb{R}^{N \times S \times F}$ to match the original shape of the multi-agent trajectories. The MLP decoder enables simultaneous non-autoregressive generation of trajectories for all aircraft, which effectively mitigates the error accumulation that is typically observed in autoregressive models. 

\section{Experimental Setup}\label{setup}
\subsection{Dataset Description} We evaluated the performance of the proposed method using ADS-B flight trajectory dataset obtained from FlightRadar24\footnote{https://www.flightradar24.com/}. The dataset spatially covers the terminal airspace within a radius of 70 nautical miles (NM) of Incheon International Airport (ICN) in South Korea from January 2023 to May 2023 (five months). Each flight trajectory data includes a timestamp, aircraft type, a sequence of geographical positions (latitude, longitude, and altitude), and other operationally relevant information such as the departure and arrival airports. In this work, we focus exclusively on arrival flights as arrival operations at ICN are considerably more complex than departures, which makes multi-agent interactions more critical. Notably, arrival and departure traffic flows are spatially segregated via dedicated flight procedures and are typically managed by different ATCs, which minimizes their interaction. Therefore, restricting the dataset to arrivals does not compromise the validity of our analysis.

To address the irregular sampling rate of ADS-B data, we resampled all trajectories at a uniform 6-second interval using the piecewise cubic Hermite interpolating polynomial (PCHIP)~\cite{fritsch1980monotone}. The resampling interval of 6 seconds was selected as a proxy for the average update rate of approximately 5 seconds in the original ADS-B data~\cite{flightradar24_historic_data}. Finer temporal resolutions result in excessively long sequences with limited performance gains, while coarser resolutions fail to capture the short-term spatio-temporal dynamics of aircraft movement. The trajectory dataset was then divided into training $\mathcal{D}_\text{train}$, validation $\mathcal{D}_\text{val}$, and test $\mathcal{D}_\text{test}$ sets using an 8:1:1 ratio (the first 80\% of samples was used for training, the next 10\% was used for validation, and the final 10\% was used for testing).

\begin{figure}[t!]			
	\centering
	\includegraphics[width=\linewidth]{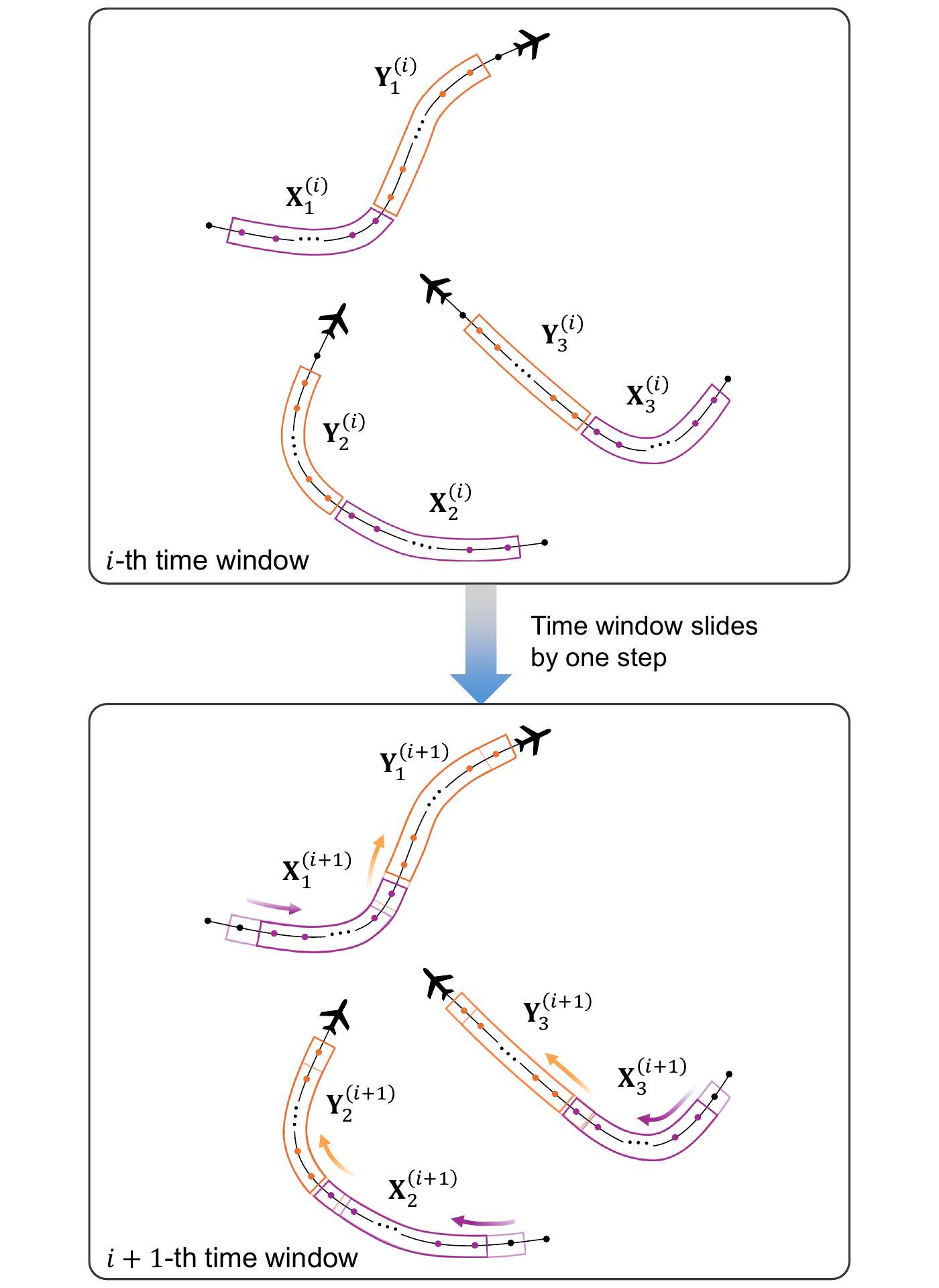}
 \vspace{-6mm} 
    \caption{Illustration of air traffic scene construction using a sliding time window.}
    \label{wdw}
 \end{figure}

For each data split, we constructed a past air traffic scene $\mathbf{X}$ and a future air traffic scene $\mathbf{Y}$ over 2-minute intervals (i.e., 20 time steps at 6-second intervals). This was done by sliding a window with one time step (6 seconds), as illustrated in Figure~\ref{wdw}. The prediction horizon of 2 minutes was motivated by operational considerations in air traffic control as it corresponds to the typical look-ahead time for short-term conflict alert (STCA) system for ATCs~\cite{eurocontrol_stca_guidelines}. Nevertheless, the proposed method is not restricted to this horizon, and other time horizons can be readily applied and should be explored in future research. To normalize the data, we applied min-max normalization to all variates using the minimum and maximum values derived from the training set. The normalization was independently applied for each variate to account for differences in their scales. As a result, we obtained a total of 509,389 traffic scenes for our experiments.

\subsection{Model and Training Configurations} MAIFormer consists of three identical stacked encoder layers. We set the latent dimension $D$ of the model to 256 with the FFN dimension of $4 \cdot D$ equating to 1024. We configured the number of attention heads $H$ to 4 in both the MMA and AA modules. The MLP decoder consists of four linear layers with sizes of 256, 128, 64, and 32, followed by a final output size of 20 (output sequence length $S$). GELU activation functions are interspersed between the linear layers, which progressively reduce the dimensionality of the latent representation to achieve the desired sequence length. The number of variates $F$ for each agent was set to 3 (latitude, longitude, and altitude). 

We trained MAIFormer using the Adam optimizer~\cite{kingma2014adam} over 300 epochs with a mini-batch size of 256. We set the initial learning rate to $10^{-4}$ and reduced it by a factor of 0.1 every 50 epochs. During training, we optimized MAIFormer using the L2 loss function (i.e., mean squared error), which computes the squared differences between predicted and actual trajectories. To mitigate overfitting, we employed early stopping and terminated the training loop if the validation error did not improve for 10 consecutive epochs. We developed and implemented MAIFormer using Python 3.9.16 and PyTorch 2.0.1~\cite{paszke2019pytorch}, while GPU acceleration was provided by CUDA 11.7. All experiments were conducted on a desktop with an Intel Core i7 processor, 32 GB of RAM, and a GeForce RTX 3060 GPU.

\subsection{Comparison Baselines} We compared the downstream performance of MAIFormer for flight trajectory prediction with that of Transformer~\cite{vaswani2017attention}, PatchTST~\cite{Yuqietal-2023-PatchTST}, and FlightBERT++~\cite{guo2024flightbert++} (single-agent methods for flight-trajectory prediction), as well as Social-LSTM~\cite{alahi2016social} and AgentFormer~\cite{yuan2021agent} (multi-agent methods). To ensure fair comparison, all models are configured with the same key hyperparameters as MAIFormer, including the latent dimension, number of layers, and feedforward network dimension. To compare the prediction accuracy, we used the mean absolute error (MAE), root mean squared error (RMSE), and mean absolute percentage error (MAPE). MAE, RMSE, and MAPE were applied separately to the latitude, longitude, and altitude dimensions. MAE provides the average magnitude of absolute prediction errors, while RMSE squares the errors before averaging and gives more weight to larger errors, which makes it sensitive to outliers. MAPE was used to evaluate relative error sizes across dimensions with different scales. These evaluation metrics are defined as follows:
\begin{equation}
\begin{aligned}
\text{MAE} &= \frac{1}{U} \frac{1}{N_u} \frac{1}{S}  \sum_{u=1}^{U}  \sum_{i=1}^{N_u} \sum_{j=1}^{S} \left| \mathbf{Y}_{u,i}^{j} - \widehat{\mathbf{Y}}_{u,i}^{j} \right| \\
\text{RMSE} &= \sqrt{ \frac{1}{U} \frac{1}{N_u} \frac{1}{S}  \sum_{u=1}^{U} \sum_{i=1}^{N_u}  \sum_{j=1}^{S} \left( \mathbf{Y}_{u,i}^{j} - \widehat{\mathbf{Y}}_{u,i}^{j} \right)^2 } \\
\text{MAPE} &= 100\% \times \frac{1}{U} \frac{1}{N_u} \frac{1}{S}  \sum_{u=1}^{U}  \sum_{i=1}^{N_u} \sum_{j=1}^{S} \left| \frac{ \mathbf{Y}_{u,i}^{j} - \widehat{\mathbf{Y}}_{u,i}^{j} }{ \mathbf{Y}_{u,i}^{j} } \right|
\end{aligned}
\end{equation}
where $U$ is the total number of air traffic scenes in the test dataset. For each traffic scene $u\in (1,2,\ldots,U)$, $N_u$ is the number of agents present in the scene, and $S$ is the prediction horizon, which we set to 20. $\mathbf{Y}_{u,i}^{j}$ is the ground truth position of the $i$-th agent at the $j$-th future time step in the $u$-th scene, and $\widehat{\mathbf{Y}}_{u,i}^{j}$ is the corresponding predicted position.

\begin{table*}[t!]
\centering
\caption{Flight trajectory prediction performance evaluation. The best performance is highlighted in \textbf{bold}, and the second-best performance is \underline{underlined}.}
\resizebox{\textwidth}{!}{ 
\begin{tabular}{c|c|ccc|ccc|ccc}
\toprule 
\multirow{2}{*}{Model} & \multirow{2}{*}{Horizon} & \multicolumn{3}{c|}{MAE\ $\downarrow$} & \multicolumn{3}{c|}{RMSE\ $\downarrow$}  & \multicolumn{3}{c}{MAPE (\%)\ $\downarrow$}\\
\cmidrule(lr){3-5} \cmidrule(lr){6-8} \cmidrule(lr){9-11}  
 & & Lat ($^{\circ}$) & Lon ($^{\circ}$) & Alt (ft) & Lat ($^{\circ}$) & Lon ($^{\circ}$) & Alt (ft)  & Lat& Lon & Alt \\

\midrule
\multirow{5}{*}{Transformer~\cite{vaswani2017attention}} 

& 1 & \cellcolor{gray!20}\textbf{0.0005} & \underline{0.0009} & \underline{27.8757} & \cellcolor{gray!20}\textbf{0.0006} & \underline{0.0009} & \underline{37.8157}  & \cellcolor{gray!20}\textbf{0.0014} & \underline{0.0007} & \underline{0.3842} \\

 & 5 & \underline{0.0012} & \cellcolor{gray!20}\textbf{0.0017} & 71.1804 & \cellcolor{gray!20}\textbf{0.0013} & \underline{0.0022} & \underline{83.5419} & 0.0035 & 0.0016 & 0.9990\\
 
 & 10 & 0.0025 & \underline{0.0032} & 149.3321 & \underline{0.0031} & \underline{0.0039} & 180.3056  & 0.0067 & \cellcolor{gray!20}\textbf{0.0025} & 2.1641 \\
 
 & 15 & 0.0045 & 0.0054 & 234.2570 & 0.0057 & 0.0067 & 285.3934  & 0.0120 & 0.0042 & 3.5570  \\
 
 & 20 & 0.0069 & 0.0080 & 319.5657 & 0.0089 & 0.0102 & 389.8004  & 0.0187 & 0.0063 & 5.1394 \\

\midrule

\multirow{5}{*}{PatchTST~\cite{Yuqietal-2023-PatchTST}} 

& 1 & \underline{0.0006} & 0.0010 & 28.8594 & \cellcolor{gray!20}\textbf{0.0006} & 0.0012 & 39.6791  & 0.0016 & 0.0007 & 0.3888 \\

 & 5 & 0.0012 & \underline{0.0018} & 86.0634 & \underline{0.0014} & 0.0024 & 99.0987 & 0.0034 & 0.0018 & 0.9998\\
 
 & 10 & 0.0024 & 0.0035 & 132.4552 & 0.0046 & 0.0055 & 177.6993  & 0.0063 & 0.0029 & 2.0001 \\
 
 & 15 & 0.0043 & 0.0049 & 223.9588 & 0.0054 & 0.0062 & 274.0511  & 0.0111 & 0.0040 & 3.3112  \\
 
 & 20 & 0.0065 & 0.0074 & 311.6529 & 0.0081 & 0.0098 & 367.2394  & 0.0161 & 0.0058 & 4.7941 \\

\midrule

\multirow{5}{*}{FlightBERT++~\cite{guo2024flightbert++}} 

& 1 & \underline{0.0006} & \cellcolor{gray!20}\textbf{0.0008} & \cellcolor{gray!20}\textbf{25.3942} & 0.0016 & 0.0017 & 74.0996  & 0.0016 & \cellcolor{gray!20}\textbf{0.0005} & \cellcolor{gray!20}\textbf{0.3592} \\

 & 5 & \cellcolor{gray!20}\textbf{0.0011} & \cellcolor{gray!20}\textbf{0.0017} & \underline{66.2858} & 0.0031 & \cellcolor{gray!20}\textbf{0.0021} & 96.1998 & \cellcolor{gray!20}\textbf{0.0031} & \cellcolor{gray!20}\textbf{0.0009} & \cellcolor{gray!20}\textbf{0.5460}\\
 
 & 10 & \underline{0.0023} & \underline{0.0032} & \underline{129.7611} & 0.0035 & \cellcolor{gray!20}\textbf{0.0027} & \underline{159.3751}  & \cellcolor{gray!20}\textbf{0.0053} & \cellcolor{gray!20}\textbf{0.0025} & \underline{1.7976} \\
 
 & 15 & 0.0041 & 0.0048 & 219.2326 & 0.0052 & 0.0062 & 271.8335  & 0.0104 & 0.0039 & 3.1479  \\
 
 & 20 & 0.0052 & 0.0059 & 287.8335 & 0.0071 & 0.0085 & 327.7224  & 0.0138 & 0.0044 & 3.6986 \\

\midrule

\multirow{5}{*}{Social-LSTM~\cite{alahi2016social}} 

& 1 & 0.0007 & 0.0009 & 32.0022 & 0.0007 & \cellcolor{gray!20}\textbf{0.0008} & 55.8011  & 0.0019 & 0.0021 & 0.4227 \\

 & 5 & 0.0017 & 0.0023 & 152.9744 & 0.0033 & 0.0034 & 181.6877 & 0.0059 & 0.0020 & 2.0886\\
 
 & 10 & 0.0038 & 0.0049 & 195.6468 & 0.0051 & 0.0055 & 251.9464  & 0.0092 & 0.0035 & 2.7888 \\
 
 & 15 & 0.0045 & 0.0054 & 234.2570 & 0.0057 & 0.0067 & 285.3934  & 0.0120 & 0.0042 & 3.5570  \\
 
 & 20 & 0.0048 & 0.0055 & 256.9070 & 0.0066 & 0.0078 & 306.6215  & 0.0122 & 0.0044 & 3.4111 \\

\midrule

\multirow{5}{*}{AgentFormer~\cite{yuan2021agent}} 
& 1 & 0.0008 & 0.0015 & 41.6010 & \underline{0.0008} & \underline{0.0015} & 42.3061 & \underline{0.0023} & 0.0012 & 0.5828 \\

& 5 & 0.0024 & 0.0031 & 139.2278 & \underline{0.0030} & 0.0037& 168.2365 & 0.0066 & 0.0024 & 2.0389 \\

& 10 & 0.0033 & 0.0041 & 187.4114 & 0.0044 & 0.0052 & 237.8679 & 0.0090 & \underline{0.0032} & 2.8028 \\

& 15 & \underline{0.0034} & \underline{0.0042} & \underline{193.7914} & \underline{0.0048} & \underline{0.0057} & \underline{257.6581} & \underline{0.0092} & \cellcolor{gray!20}\textbf{0.0033} & \underline{2.8636} \\

& 20 & \underline{0.0039} & \underline{0.0054} & \underline{219.3697} & \underline{0.0056} & \underline{0.0066} & \underline{292.7384} & \underline{0.0107} & \underline{0.0043} & \underline{3.2515} \\

\midrule
\multirow{5}{*}{MAIFormer} 
& 1 & 0.0008 & 0.0014 & 29.1527& \underline{0.0008} & 0.0016 & \cellcolor{gray!20}\textbf{33.5542} & \underline{0.0023} & 0.0011 & 0.4104 \\

& 5 & \underline{0.0012} & 0.0020 & \cellcolor{gray!20}\textbf{60.1332} & \underline{0.0014} & \cellcolor{gray!20}\textbf{0.0021} & \cellcolor{gray!20}\textbf{67.7342} & \underline{0.0033} & \underline{0.0015} &  \underline{0.8347} \\

& 10 & \cellcolor{gray!20}\textbf{0.0022} & \cellcolor{gray!20}\textbf{0.0030} & \cellcolor{gray!20}\textbf{99.7254} & \cellcolor{gray!20}\textbf{0.0023} & \underline{0.0034} & \cellcolor{gray!20}\textbf{115.0246} & \underline{0.0056} & \cellcolor{gray!20}\textbf{0.0025} & \cellcolor{gray!20}\textbf{1.4943} \\

& 15 & \cellcolor{gray!20}\textbf{0.0028} & \cellcolor{gray!20}\textbf{0.0041} & \cellcolor{gray!20}\textbf{131.7546} & \cellcolor{gray!20}\textbf{0.0033} & \cellcolor{gray!20}\textbf{0.0048} & \cellcolor{gray!20}\textbf{159.4441} & \cellcolor{gray!20}\textbf{0.0078} & \underline{0.0034} & \cellcolor{gray!20}\textbf{2.1112} \\

& 20 & \cellcolor{gray!20}\textbf{0.0037} & \cellcolor{gray!20}\textbf{0.0047} & \cellcolor{gray!20}\textbf{164.1826} & \cellcolor{gray!20}\textbf{0.0046} & \cellcolor{gray!20}\textbf{0.0065} & \cellcolor{gray!20}\textbf{198.2483} & \cellcolor{gray!20}\textbf{0.0101} & \cellcolor{gray!20}\textbf{0.0037} & \cellcolor{gray!20}\textbf{2.7827} \\

\midrule
{PI (\%)} & - & \textbf{21.11} & \textbf{9.81} & \textbf{34.88} & \textbf{32.64} & \textbf{13.93} & \textbf{38.07}  & \textbf{16.67} & \textbf{3.66} & \textbf{22.08}

\\
\bottomrule 
\end{tabular}
} 
\label{error_table}
\end{table*}

\section{Numerical Experiments}\label{result} 

\subsection{Quantitative Analysis}
Table~\ref{error_table} shows the flight trajectory prediction performance of MAIFormer and all baseline models. To enable more comprehensive analysis of multi-horizon flight trajectory prediction, we report prediction errors across time horizons of 1, 5, 10, 15, and 20 steps, which correspond to 6 seconds, 30 seconds, 1 minute, 1.5 minutes, and 2 minutes into the future. We also computed the performance improvement (PI) ratio as $\left\vert p_2 - p_1 \right\vert / p_2$ to compare the best performance $p_1$ to the second-best performance $p_2$. We computed the PI ratios based on the average across all time steps of each model.

\begin{figure*}[t!]			
\centering
\includegraphics[width=\textwidth]{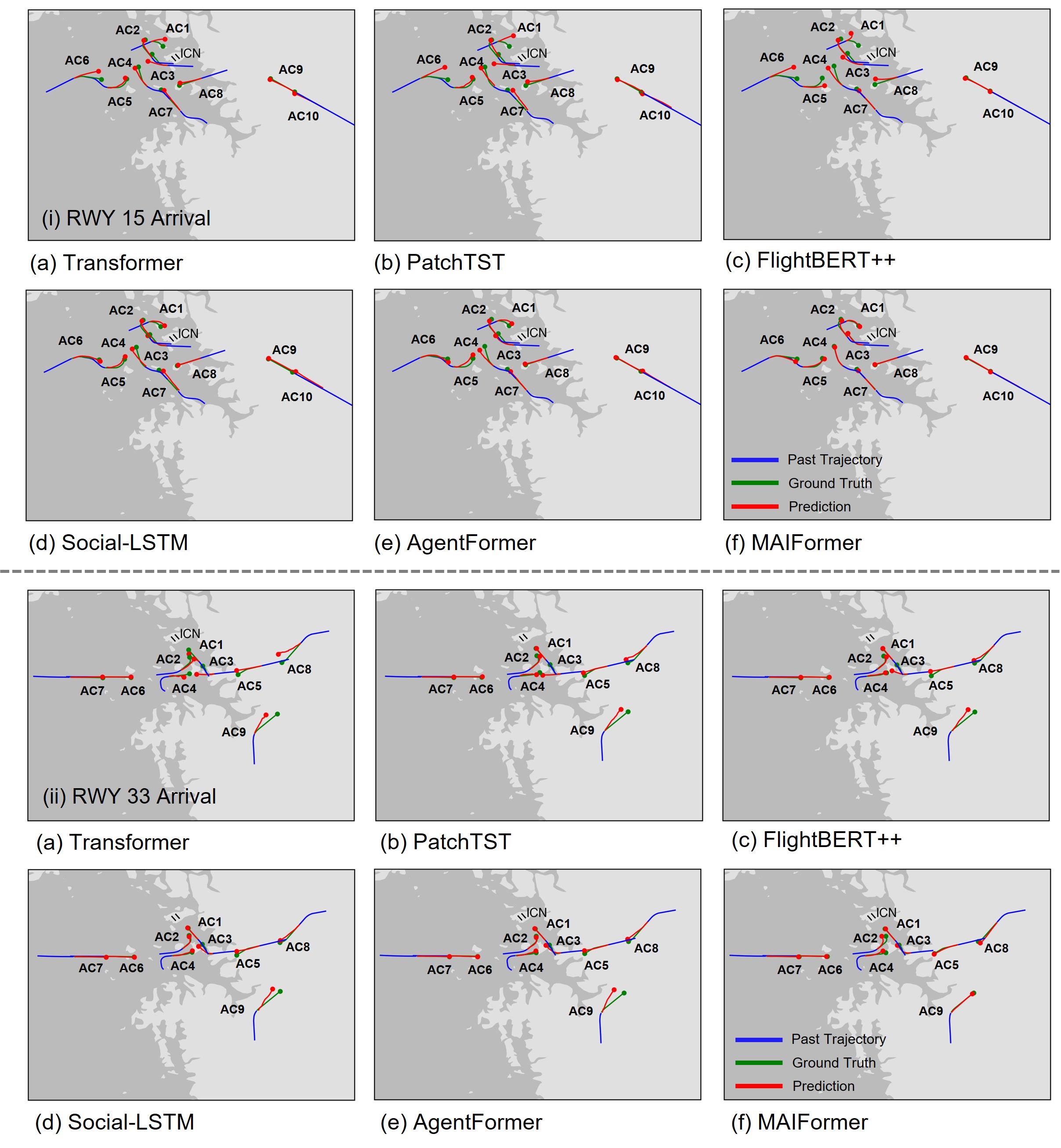}
\vspace{-5mm}
\caption{Flight trajectory prediction results for two illustrative flight scenarios. Each scenario is presented in two rows showing predictions from six different models.}
\label{prediction}
\end{figure*}

MAIFormer substantially outperformed all baseline models. Compared to the second-best model, the average MAE of MAIFormer showed reductions in the latitude, longitude, and altitude dimensions of 21.11\%, 9.81\%, and 34.88\%, respectively. Improvements in RMSE were even more pronounced, with error reductions of 32.64\%, 13.93\%, and 38.07\% in the corresponding dimensions. We attribute these stronger RMSE gains to MAIFormer’s improved ability to capture the underlying dynamics of air traffic scenes. MAIFormer also achieved MAPE reductions of 16.67\%, 3.66\%, and 22.08\% in latitude, longitude, and altitude dimensions, respectively.

When analyzing prediction errors for different prediction horizons, the single-agent models (Transformer, PatchTST, and FlightBERT++) slightly outperformed the multi-agent models Social-LSTM, AgentFormer, and MAIFormer at early time steps (e.g., horizons 1 and 5). This behavior is likely due to short-term future flight trajectories being largely dominated by the dynamics of individual aircraft, such as current velocity and heading, which means that they can be extrapolated easily without explicitly modeling interactions with surrounding traffic. In contrast, longer-term future trajectories (e.g., horizons 10, 15, and 20) are more strongly influenced by surrounding aircraft and overall traffic patterns. As a result, single-agent models generally underperform in longer-term predictions due to their inability to account for the contextual interactions between traffic. Notably, compared with other multi-agent models (Social-LSTM and AgentFormer), MAIFormer achieves more consistent and substantial performance gains over single-agent models across all prediction horizons.

\subsection{Qualitative Analysis}
For qualitative evaluation, we visualized the flight trajectory prediction results for two illustrative scenarios in Figure~\ref{prediction}, which correspond to arrivals at ICN Runway 15, and Runway 33. In each plot, the blue line represents the past trajectory (used as input to the models), the green line indicates the ground-truth future trajectory, and the red line represents the model’s prediction. Notably, in the Runway 15 arrival scenario (top two rows of Figure~\ref{prediction}), more complex and irregular traffic patterns occur near the airport compared with the Runway 33 scenario. This is due to the northern part of the ICN airspace being largely unavailable due to its proximity to the North Korean border, which requires arriving traffic to merge earlier than in the Runway 33 scenario.

For the Runway 15 scenario, single-agent models struggle more to accurately predict trajectories for aircraft near the runway compared to multi-agent models. In the vicinity of the runway, inter-aircraft separation becomes small, which makes capturing aircraft interactions crucial. The multi-agent models, Social-LSTM and AgentFormer, perform reasonably well even in the dense area near the runway, but their predicted trajectories for AC4 and AC5 are unrealistically close to each other. In contrast, MAIFormer consistently produces accurate predictions for all aircraft in this region, and its predicted trajectories maintain appropriate separation. For aircraft that are farther from the runway, both single-agent and multi-agent models generate relatively accurate predictions.

For the Runway 33 arrival scenario (bottom two rows of Figure~\ref{prediction}), all baseline models perform reasonably well overall but still produce inaccurate predictions for several aircraft. For example, the baseline models fail to capture AC9’s future trajectory and incorrectly predict that it will arrive ahead of AC8, which is already established within a structured arrival flow. In contrast, MAIFormer successfully captures the subtle interactions between these two aircraft and correctly places AC9 behind AC8 in the arrival sequence. This example highlights MAIFormer’s stronger capability of understanding traffic scenes and modeling inter-agent interactions.

\begin{figure}[t!]			
\centering
\includegraphics[width=\linewidth]{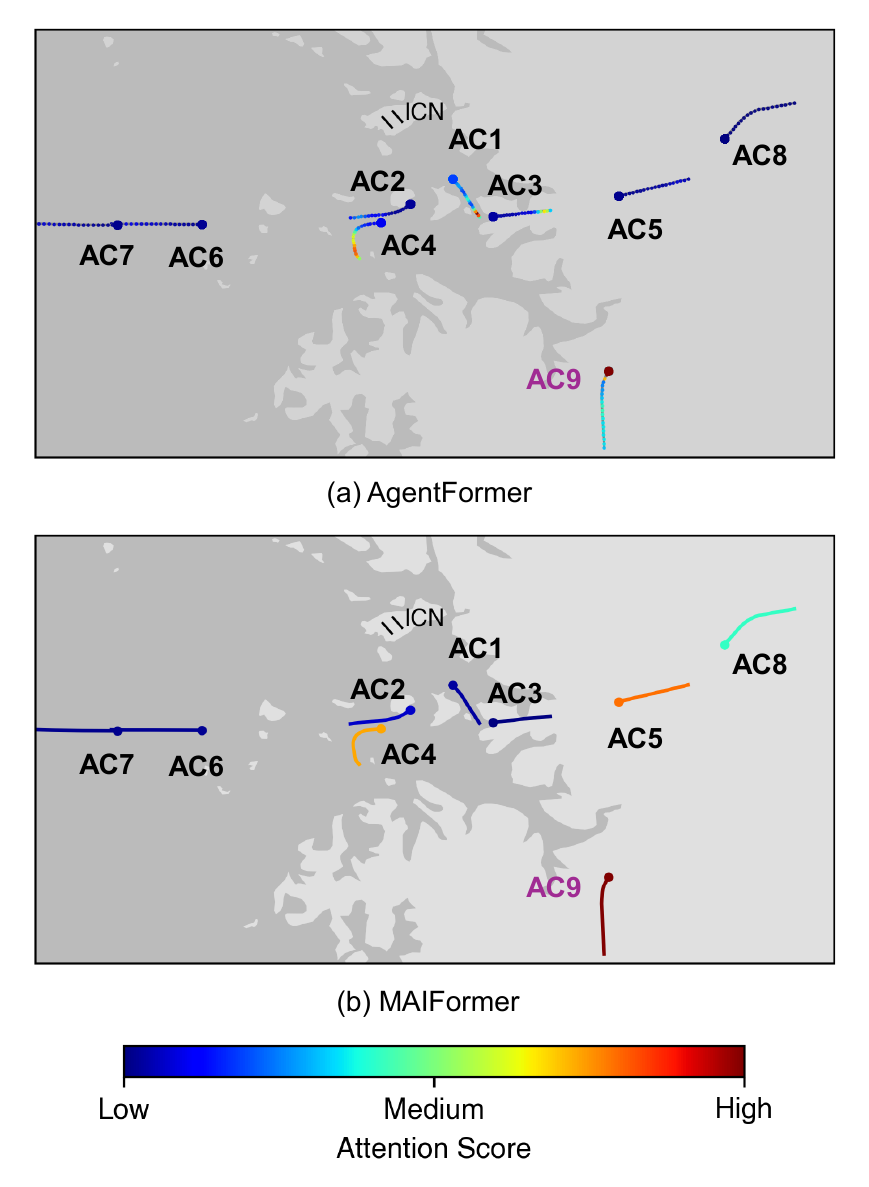}
\vspace{-7mm}
\caption{Visualization of attention scores for the Runway 33 arrival scenario in Figure~\ref{prediction}. The color bar represents the magnitude of attention. Red indicates higher attention scores, and blue indicates lower scores.}
\label{attens}
\end{figure} 

Figure~\ref{attens} illustrates the attention distribution across different aircraft computed by AgentFormer and MAIFormer for the Runway 33 arrival scenario (Figure~\ref{prediction}, bottom row). The attention scores were visualized from the perspective of AC9 (which serves as the query), and the color reflects how much attention it assigns to each surrounding aircraft. AgentFormer computes attention scores at the trajectory point level and models the relationship between individual points across different time steps. As a result, the attention score can vary even within the same aircraft’s trajectory. This results in fine-grained information, but the excessively detailed attention distributions are inherently difficult for human operators to interpret. As the number of aircraft in a traffic scene increases, the complexity of interpreting these point-to-point relationships grows rapidly, which reduces their practical utility in operational contexts.

In contrast, MAIFormer computes attention at the agent level and assigns a single attention score to each aircraft. These agent-to-agent attention values are far more interpretable as they represent the relative importance of each aircraft as perceived by the querying aircraft (AC9 in this case). This abstraction enables a more intuitive and human-understandable interpretation of inter-agent relationships, which is particularly valuable in safety-critical domains such as ATM. Interestingly, MAIFormer assigns high attention not only to AC5 and AC8 but also to AC4. A possible explanation for this behavior is that AC9 may have had the potential to enter the arrival sequence earlier before AC5, depending on the location and speed of AC4. The model appears to have incorporated about this possibility and thus assigned a relatively high attention weight to AC4.

\subsection{Attention Entropy Analysis}
We performed an attention entropy analysis to quantitatively evaluate and compare the interpretability of the agent-level attention patterns produced by MAIFormer and AgentFormer. For each query agent $i$, we computed the Shannon entropy of its attention distribution over all key aircraft $j$, which provides a scalar measure of how concentrated or diffused the learned interactions are. The attention entropy is defined as follows:
\begin{equation}
    H = \frac{1}{Q}\sum_{i=1}^{Q} \left(-\sum_{j=1}^{K} a_{ij} \log(a_{ij}+\epsilon)\right)
\end{equation}
where $Q$ and $K$ are the numbers of query and key tokens, respectively, $a_{ij}$ is the normalized attention weight assigned from query agent $i$ to key agent $j$, and $\epsilon$ is a small constant that was added for numerical stability. Lower entropy indicates that the attention is concentrated on a smaller subset of agents, while higher entropy indicates a more diffuse attention distribution. We report the attention entropy as a function of the number of aircraft in the traffic scene to examine how attention concentration varies with traffic complexity.
\begin{figure}[t!]			
\centering
\includegraphics[width=\linewidth]{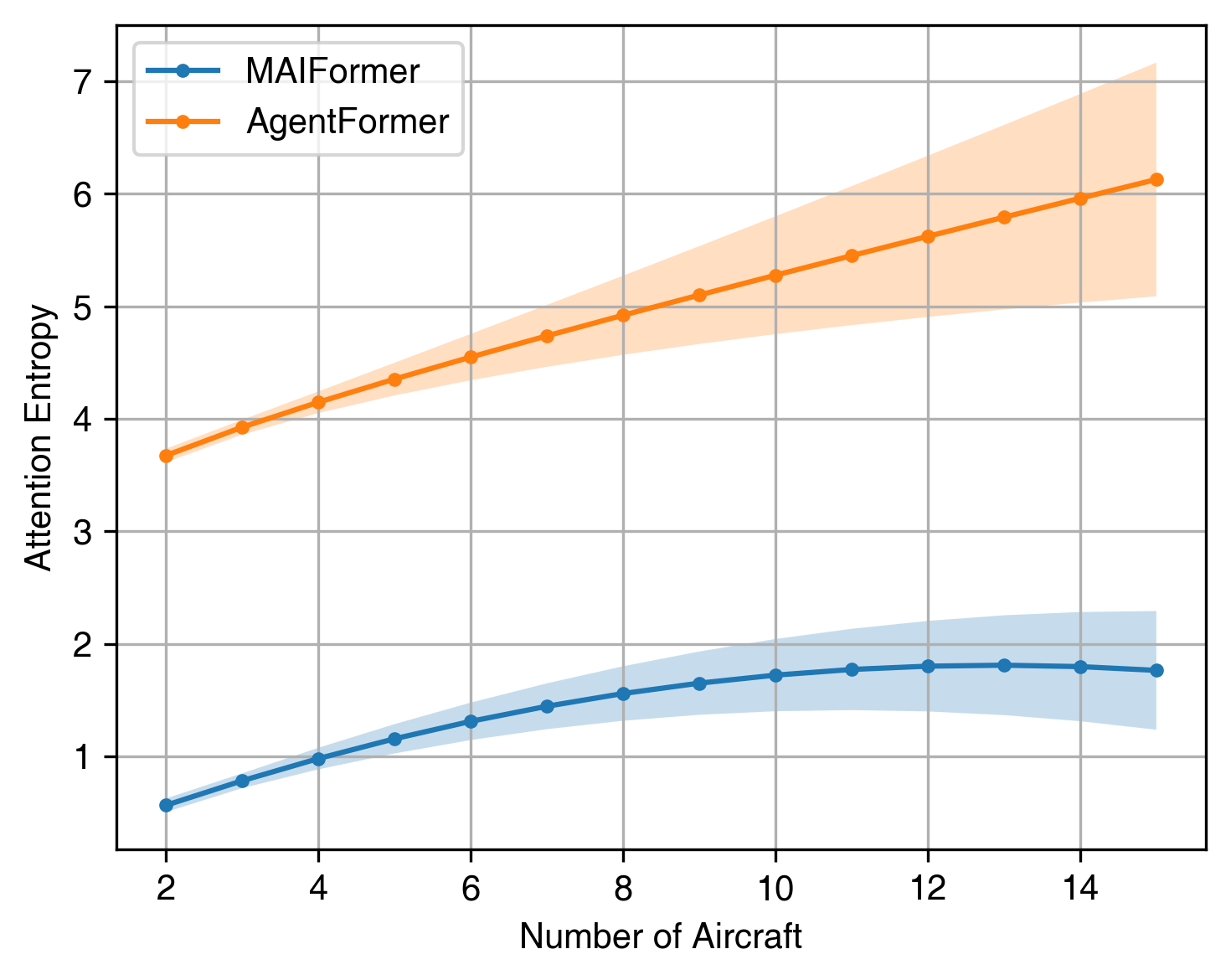}
\vspace{-7mm}
\caption{Comparison of attention entropy with respect to the number of aircraft.}
\label{entropy}
\end{figure}

\begin{table*}[b!]
\centering
\caption{Ablation studies. The best performance is highlighted in \textbf{bold}, and the second-best performance is \underline{underlined}.}
\resizebox{\textwidth}{!}{ 
\begin{tabular}{c|c|ccc|ccc|ccc}
\toprule 
\multirow{2}{*}{Model} & \multirow{2}{*}{Horizon} & \multicolumn{3}{c|}{MAE\ $\downarrow$} & \multicolumn{3}{c|}{RMSE\ $\downarrow$}  & \multicolumn{3}{c}{MAPE (\%)\ $\downarrow$}\\
\cmidrule(lr){3-5} \cmidrule(lr){6-8} \cmidrule(lr){9-11}  
 & & Lat ($^{\circ}$) & Lon ($^{\circ}$) & Alt (ft) & Lat ($^{\circ}$) & Lon ($^{\circ}$) & Alt (ft)  & Lat& Lon & Alt \\

\midrule
\multirow{5}{*}{\makecell{MAIFormer\\(w/o MMA)}}

& 1  & \underline{0.0009} & \underline{0.0015} &  \underline{33.2131} & \underline{0.0010} & \cellcolor{gray!20}\textbf{0.0016} &  \underline{36.4879} &  \underline{0.0024} &  \underline{0.0012} &  \underline{0.4211} \\

& 5  & \underline{0.0018} & \underline{0.0024} & \underline{95.6454} & 0.0029 & 0.0033 & \underline{132.4674} & 0.0041 & 0.0023 & 1.4441 \\

& 10 & 0.0031 & 0.0040 & 155.4566 & 0.0040 & 0.0051 & 171.6697 & 0.0088 & 0.0035 & 2.7714 \\

& 15 & \underline{0.0039} & 0.0048 & \underline{196.6767} & 0.0049 & \underline{0.0056} & 282.4156 & \underline{0.0095} & \underline{0.0038} & \underline{2.8912} \\

& 20 & \underline{0.0042} & \underline{0.0056} & \underline{225.7871} & \underline{0.0052} & \underline{0.0069} & \underline{299.1411} & \underline{0.0115} & \underline{0.0047}& \underline{3.4396} \\

\midrule

\multirow{5}{*}{\makecell{MAIFormer\\(w/o AA)}}

& 1  & \underline{0.0009} & \underline{0.0015} & 37.4491 & \underline{0.0010} & 0.0017 & 43.1547 & 0.0025 & 0.0013 & 0.4227 \\

& 5  & \underline{0.0018} & 0.0026 & 119.0749 & \underline{0.0028} & \underline{0.0032} & 141.6576 & \underline{0.0040} & \underline{0.0022} & \underline{1.4174} \\

& 10 & \underline{0.0029} & \underline{0.0038} & \underline{141.4881} & \underline{0.0038} & \underline{0.0048} & \underline{166.6623} & \underline{0.0085} & \underline{0.0034} & \underline{1.5551} \\

& 15 & \underline{0.0039} & \underline{0.0047} & 201.7765 & \underline{0.0048} & 0.0058 & \underline{275.4671} & 0.0108 & 0.0039 & 2.8991 \\

& 20 & 0.0043 & 0.0057 & 244.5641 & 0.0055 & 0.0070 & 301.1086 & 0.0116 & 0.0048 & 3.4912 \\
\midrule

\multirow{5}{*}{\makecell{MAIFormer\\(w/o inverting)}}

& 1  & 0.0011 & 0.0017 & 36.2344 & 0.0011 & 0.0018 & 45.1473 & 0.0028 & 0.0014 & 0.4474 \\

& 5  & 0.0024 & 0.0030 & 142.1475 & 0.0032 & 0.0041 & 181.7415 & 0.0064 & 0.0031 & 2.3477 \\

& 10 & 0.0036 & 0.0045 & 200.5464 & 0.0044 & 0.0058 & 264.1141 & 0.0121 & 0.0043 & 2.9991 \\

& 15 & 0.0041 & 0.0052 & 221.4414 & 0.0052 & 0.0061 & 281.7987 & 0.0137 & 0.0045 & 3.0041 \\

& 20 & 0.0049 & 0.0058 & 264.4875 & 0.0061 & 0.0074 & 322.1772 & 0.0145 & 0.0049 & 3.6468 \\

\midrule

\multirow{5}{*}{\makecell{MAIFormer\\(Proposed)}}

& 1 & \cellcolor{gray!20}\textbf{0.0008} & \cellcolor{gray!20}\textbf{0.0014} & \cellcolor{gray!20}\textbf{29.1527} & \cellcolor{gray!20}\textbf{0.0008} & \cellcolor{gray!20}\textbf{0.0016} & \cellcolor{gray!20}\textbf{33.5542} & \cellcolor{gray!20}\textbf{0.0023} & \cellcolor{gray!20}\textbf{0.0011} & \cellcolor{gray!20}\textbf{0.4104} \\

& 5 & \cellcolor{gray!20}\textbf{0.0012} & \cellcolor{gray!20}\textbf{0.0020} & \cellcolor{gray!20}\textbf{60.1332} & \cellcolor{gray!20}\textbf{0.0014} & \cellcolor{gray!20}\textbf{0.0021} & \cellcolor{gray!20}\textbf{67.7342} & \cellcolor{gray!20}\textbf{0.0033} & \cellcolor{gray!20}\textbf{0.0015} & \cellcolor{gray!20}\textbf{0.8347} \\

& 10 & \cellcolor{gray!20}\textbf{0.0022} & \cellcolor{gray!20}\textbf{0.0030} & \cellcolor{gray!20}\textbf{99.7254} & \cellcolor{gray!20}\textbf{0.0023} & \cellcolor{gray!20}\textbf{0.0034} & \cellcolor{gray!20}\textbf{115.0246} & \cellcolor{gray!20}\textbf{0.0056} & \cellcolor{gray!20}\textbf{0.0025} & \cellcolor{gray!20}\textbf{1.4943} \\

& 15 & \cellcolor{gray!20}\textbf{0.0028} & \cellcolor{gray!20}\textbf{0.0041} & \cellcolor{gray!20}\textbf{131.7546} & \cellcolor{gray!20}\textbf{0.0033} & \cellcolor{gray!20}\textbf{0.0048} & \cellcolor{gray!20}\textbf{159.4441} & \cellcolor{gray!20}\textbf{0.0078} & \cellcolor{gray!20}\textbf{0.0034} & \cellcolor{gray!20}\textbf{2.1112} \\

& 20 & \cellcolor{gray!20}\textbf{0.0037} & \cellcolor{gray!20}\textbf{0.0047} & \cellcolor{gray!20}\textbf{164.1826} & \cellcolor{gray!20}\textbf{0.0046} & \cellcolor{gray!20}\textbf{0.0065} & \cellcolor{gray!20}\textbf{198.2483} & \cellcolor{gray!20}\textbf{0.0101} & \cellcolor{gray!20}\textbf{0.0037} & \cellcolor{gray!20}\textbf{2.7827} \\
\bottomrule 
\end{tabular}
} 
\label{ablation}
\end{table*}

As shown in Figure~\ref{entropy}, as the number of aircraft grows, the attention entropy of AgentFormer increases substantially and remains relatively high even for small traffic scenes. In contrast, MAIFormer consistently exhibits lower attention entropy and a much slower growth rate with increasing traffic density, which indicates that the agent-level attention patterns are more concentrated and interpretable. Notably, the entropy curve of MAIFormer gradually converges, which suggests that even as the traffic density increases, MAIFormer does not diffuse its focus but instead attends to a limited set of key aircraft that most strongly influences the future trajectory of the query aircraft. Such focused attention patterns and stable entropy growth are desirable as they improve the usability of MAIFormer for interaction-aware decision support and analysis of interaction priorities in historical traffic data.

\subsection{Ablation Studies}
We also conducted ablation studies to examine the contributions of key components in MAIFormer. We removed the MMA module and allowed the AA module to capture both spatio-temporal dynamics and inter-agent interactions to assess whether aggregated agent tokens alone are sufficient for accurate prediction. We then retained MMA but removed AA and the masking strategy in MMA to evaluate how agent attention over variate tokens affects performance. Finally, we replaced the inverted embedding with a standard temporal embedding while keeping the rest of the MAIFormer architecture unchanged to quantify the impact of the embedding strategy.

Table~\ref{ablation} summarizes the ablation results. All ablated variants exhibit poorer performance compared to the full MAIFormer architecture. In particular, when either MMA or AA is removed, the model is still able to produce reasonable predictions but consistently shows greater prediction errors than the complete model. This suggests that while MMA or AA alone can partially capture relevant information, separating spatio-temporal dynamics and inter-agent interactions into dedicated hierarchical attention modules leads to more effective representation learning and better performance. Moreover, replacing the inverted embedding with standard temporal embedding results in the largest performance degradation. This finding indicates that conventional temporal embeddings are insufficient to capture complex and airspace-specific flight patterns, and the combination of the inverted embedding and hierarchical attention is essential for achieving the best prediction accuracy with MAIFormer.

\section{Conclusion}\label{conclusion}
We have proposed MAIFormer as a multi-agent flight-trajectory prediction model that learns both the spatio-temporal patterns of individual aircraft and the interaction patterns between them. Its effectiveness was demonstrated using real-world air traffic surveillance data, and the experimental results showed that it achieves better prediction accuracy compared to baseline models. Its hierarchical attention architecture allows it to produce interpretable attention distributions that explicitly indicate the degree of influence that each surrounding aircraft has on a given aircraft’s future trajectory. 

As a potential avenue for practical deployment and future extension, the learned agent-level attention scores can be valuable for a broader range of ATC-related applications. For example, in the development of autonomous ATC systems that aim to imitate historical controller behavior, the aircraft-level attention scores generated by the system can provide insights into how the systems prioritize and monitor specific aircraft in dynamic traffic conditions.

Nevertheless, despite its strong performance, several aspects require further exploration to improve its practical applicability in real-world air traffic operations. MAIFormer provides insights into aircraft interactions through its agent attention, but it remains uncertain whether these learned patterns truly reflect a human ATC’s attention. Future research should involve human-in-the-loop simulations or expert assessments to evaluate the interpretability and operational alignment of the attention distributions produced by the proposed method. Furthermore, the model relies solely on flight trajectory data, but in real-world operations, external contextual factors play a significant role in ATC decision making and aircraft behavior, including meteorological conditions, flight procedures, and airspace constraints. Incorporating such conditional information into the model as exogenous inputs would improve the prediction accuracy and bring the model closer to the standard required for real-world deployment in an ATM system.

\bibliographystyle{IEEEtran}
\bibliography{sample}

\begin{IEEEbiography}[{\includegraphics[width=1in,height=1.25in,clip,keepaspectratio]{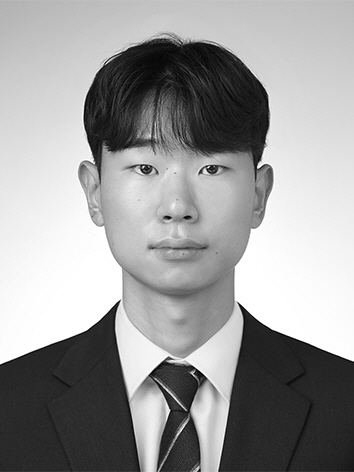}}]{Seokbin Yoon} received the B.S and M.S degrees in Air Transport, Transportation, and Logistics from Korea Aerospace University in 2022 and 2024, respectively. He is currently a research assistant at Air Transportation Modeling and Control Laboratory, Korea Aerospace University. His research interests include air traffic management, flight trajectory modeling, and deep learning.
\end{IEEEbiography}

\begin{IEEEbiography}[{\includegraphics[width=1in,height=1.25in,clip,keepaspectratio]{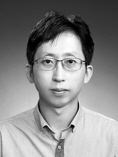}}]{Keumjin Lee} received the B.S. and M.S. degrees in aeronautical engineering from Seoul National University, South Korea, in 1999 and 2001, respectively, and the Ph.D. degree in aerospace engineering from Georgia Institute of Technology, Atlanta, in 2008. He was a Researcher with the Air Traffic Management Department, Electronic Navigation Research Institute, Japan, and also with the Department of Aviation Research, Korea Transport Institute, South Korea. He is currently a Professor with the Department of Air Transportation, Transportation, and Logistics, Korea Aerospace University, South Korea. His research interests include modeling and optimization and their applications to air traffic management and airspace design.
\end{IEEEbiography}

\end{document}